\documentclass{article}

\usepackage{arxiv}

\usepackage[utf8]{inputenc} 
\usepackage[T1]{fontenc}    
\usepackage{hyperref}       
\usepackage{url}            
\usepackage{booktabs}       
\usepackage{amsfonts}       
\usepackage{nicefrac}       
\usepackage{microtype}      
\usepackage{lipsum}
\usepackage{graphicx}

\title{NITES: A Non-Parametric Interpretable Texture Synthesis Method}

\author{
  Xuejing Lei\\
  Media Communications Lab\\
  University of Southern California\\
  Los Angeles, CA, USA \\
  \texttt{xuejing@usc.edu} \\
   \And
  Ganning Zhao\\
  Media Communications Lab\\
  University of Southern California\\
  Los Angeles, CA, USA \\
  \texttt{ganningz@usc.edu} \\
  \And
  C.-C. Jay Kuo\\
  Media Communications Lab\\
  University of Southern California\\
  Los Angeles, CA, USA \\
  \texttt{cckuo@sipi.usc.edu} \\
}

\begin{document}
\maketitle

\begin{abstract}
A non-parametric interpretable texture synthesis method, called the
NITES method, is proposed in this work. Although automatic synthesis of
visually pleasant texture can be achieved by deep neural networks
nowadays, the associated generation models are mathematically
intractable and their training demands higher computational cost. NITES
offers a new texture synthesis solution to address these shortcomings.
NITES is mathematically transparent and efficient in training and
inference. The input is a single exemplary texture image. The NITES
method crops out patches from the input and analyzes the statistical
properties of these texture patches to obtain their joint
spatial-spectral representations. Then, the probabilistic distributions
of samples in the joint spatial-spectral spaces are characterized.
Finally, numerous texture images that are visually similar to the
exemplary texture image can be generated automatically. Experimental
results are provided to show the superior quality of generated texture
images and efficiency of the proposed NITES method in terms of both
training and inference time.
\end{abstract}


\section{Introduction}

Automatic synthesis of visually pleasant texture that resembles
exemplary texture finds applications in computer graphics. Texture
synthesis has been studied for several decades since it is also of
theoretical interest in texture analysis and modeling. Texture can be
synthesized pixel-by-pixel \cite{de1997multiresolution,
efros1999texture, wei2000fast} or patch-by-patch~\cite{efros2001image,
liang2001real, cohen2003wang,kwatra2003graphcut,wu2004feature} based on
an exemplary pattern. For the pixel-based synthesis, a pixel conditioned
on its squared neighbor was synthesized using the conditional
probability and estimated by a statistical method in
\cite{efros1999texture}. Generally, patch-based texture synthesis yields
higher quality than pixel-based texture synthesis. Yet, to search the
whole image for patch-based synthesis is extremely slow
\cite{efros1999texture, liang2001real}. To speed up, one can stitch
small patches of the exemplary texture to form a larger region
~\cite{efros2001image, cohen2003wang, kwatra2005texture}. Although these
methods can produce texture of higher quality, its diversity is limited.
Besides texture synthesis in the spatial domain, one can transform
texture images from the spatial domain to the spectral domain with
certain filters (or kernels) and exploit the statistical correlation of
filter responses for texture synthesis. Commonly used kernels include
the Gabor filters~\cite{heeger1995pyramid} and the steerable pyramid
filter banks~\cite{portilla2000parametric}. 

We have witnessed amazing quality improvement of synthesized texture in
the last 5-6 years due to the resurgence of neural networks. Texture
synthesis based on deep learning (DL), such as convolutional neural
networks (CNNs) and generative adversarial networks (GANs), yield
visually pleasant results. DL-based methods learn transform kernels from
numerous training data~\cite{gatys2015texture, liu2016texture,
risser2017stable, li2017diversified, li2017universal,
ustyuzhaninov2017does, shi2020fast} through end-to-end optimization.
However, these methods have two main shortcomings: 1) lack of
mathematical transparency and 2) higher training and inference
complexity.  To address them, we investigate a non-parametric and
interpretable texture synthesis method, called NITES, in this work. 

NITES consists of three steps. First, it analyzes the exemplary texture
to obtain its joint spatial-spectral representations.  Second, the
probabilistic distributions of training samples in the joint
spatial-spectral spaces are characterized. Finally, new texture images
are generated by mimicking probabilities of source texture images. In
particular, we adopt a data-driven transform, known as the channel-wise
(c/w) Saab transform~\cite{chen2020pixelhopplus}, which provides a
powerful representation in the joint spatial-spectral space.  The c/w
Saab transform is derived from the successive subspace learning (SSL)
theory~\cite{kuo2016understanding, kuo2017cnn,
kuo2019interpretable,chen2020pixelhop}. We will show that NITES can
generate high quality texture at lower complexity. 

The rest of the paper is organized as follows. The framework of
successive subspace embedding and generation is described in Sec.
\ref{sec:SSL}. The NITES method is proposed in Sec.~\ref{sec:method}.
Experimental results are shown in Sec.~\ref{sec:experiments}. Concluding
remarks are given in Sec.~\ref{sec:conclusion}. 

\section{Successive Subspace Embedding and Generation}\label{sec:SSL}

In this section, we would like to explain a high-level idea behind the
NITES method, which is called the successive subspace embedding and
generation principle. Consider an input signal space denoted by
$\tilde{S}_0$, and a sequence of subspaces denoted by $\tilde{S}_1,
\cdots, \tilde{S}_n$. Their dimensions are denoted by $\tilde{D}_0$,
$\tilde{D}_1$, $\cdots$, $\tilde{D}_n$.  They are related with each
other by the constraint -- any element in $\tilde{S}_{i+1}$ is formed by
an affine combination of elements in $\tilde{S}_{i}$, where $i=0,
\cdots, n-1$. 

\begin{figure*}[t]
\centering
\includegraphics[width=0.9\linewidth]{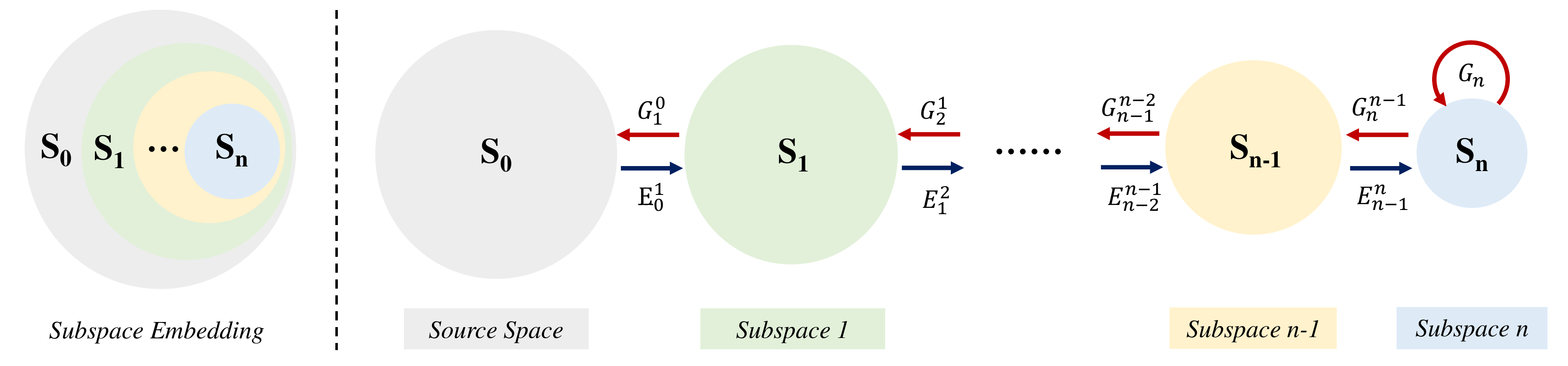}
\caption{Illustration of the concept of successive subspace embedding and
generation, where blue arrows indicates a sequence of subspace embedding
processes and red arrows indicates a sequence of subspace generation
processes.}\label{fig:subspace}
\end{figure*}

An affine transform can be converted to a linear transform by augmenting
vector $\tilde{{\bf a}}$ in $\tilde{S}_{i}$ via ${\bf a}=(\tilde{{\bf
a}}^T, 1)^T$. We use $S_i$ to denote the augmented space of
$\tilde{S}_{i}$ and $D_i=\tilde{D}_i+1$. Then, we have the following
embedding relationship
\begin{equation}\label{eq:embedding}
S_n \subset S_{n-1} \subset \cdots \subset S_1 \subset S_0,
\end{equation}
and
\begin{equation}\label{eq:dim}
D_n < D_{n-1} < \cdots <  D_1 < D_0.
\end{equation}
This concept is illustrated in Fig. \ref{fig:subspace} with $n=2$.

We use texture embedding and generation as an example to explain this
concept. To generate homogeneous texture, we collect a collection of
texture patches cropped out of exemplary texture as the input set.
Suppose that each texture patch has RGB three color channels, and a
spatial resolution $N\times N$. Then, the input set has a dimension of $3
N^2$ and its augmented space $S_0$ has a dimension of $D_0=3 N^2 + 1$.
If $N=32$, we have $D_0=3073$ which is too high to find an effective
generation model directly. 

To address this challenge, we build a sequence of embedded subspaces
$S_0$, $S_1$, $\cdots$, $S_n$ with decreasing dimensions. We call $S_0$
and $S_n$ the "source" space and the "core" subspace, respectively.
We need to find an effective embedded subspace $S_{i+1}$ from $S_i$, and
such an embedding model is denote by $E_i^{i+1}$. Proper subspace
embedding is important since it determines how to decompose an input
space into the direct sum of two subspaces in the forward embedding path.
Although we choose one of the two for further processing, we also need
to record the probabilistic relationship of the two decomposed
subspaces so that samples of diversity and fidelity can be generated in
the reverse generation path. 

In the reverse generation path, we begin with the generation of samples
in $S_n$ by studying its own statistics. This is accomplished by
generation model $G_n$. Then, conditioned on a generated sample in
$S_{i+1}$, we generate a new sample in $S_{i}$ through a generation
model denoted by $G_{i+1}^i$.  In Fig. \ref{fig:subspace}, we use blue
and red arrows to indicate a sequence of subspace embedding and
generation processes, respectively. This idea can be implemented as a
non-parametric method since we can choose subspaces $S_1$, $\cdots$,
$S_n$, flexibly in a feedforward manner. One specific design is
elaborated in the next section. 

\section{Proposed NITES Method}\label{sec:method}

The proposed NITES method is presented in this section. To begin with,
we provide a system overview of the NITES method in
Sec.~\ref{subsec:system}. Next, we discuss the successive subspace
embedding (SSE) scheme based on the c/w Saab transform in
Sec.~\ref{subsec:sse}. After that, we examine the problem of core
subspace generation (CSG) in Sec.~\ref{subsec:csg}.  Finally, we
describe the successive subspace generation (SSG) process in
Sec.~\ref{subsec:ssg}. 

\subsection{System Overview}\label{subsec:system}

\begin{figure*}[tb]
\centering
\includegraphics[width=1.0\linewidth]{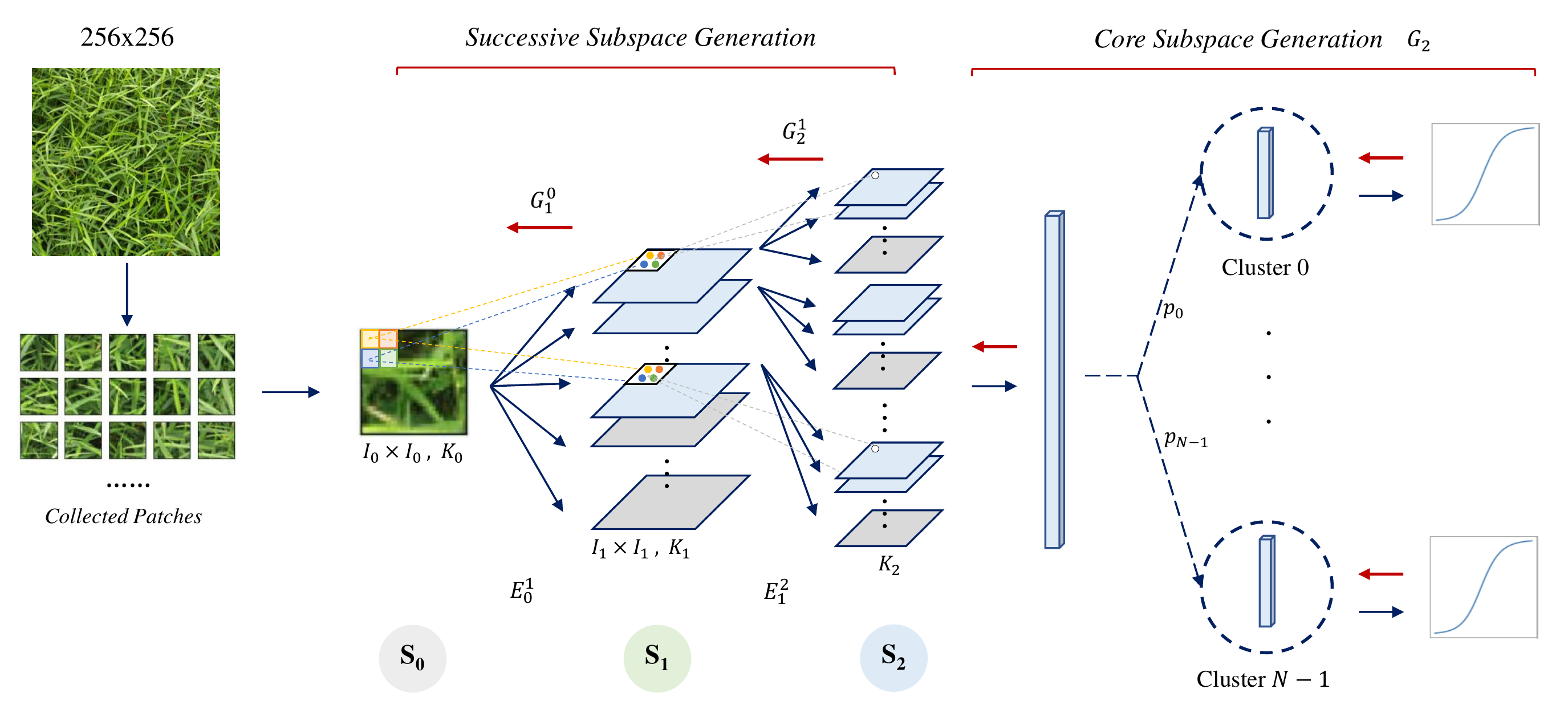}
\caption{An overview of the proposed NITES method.}\label{fig:framework}
\end{figure*}

An overview of the NITES method is given in Fig.~\ref{fig:framework}.
The exemplary color texture image has a spatial resolution of $256
\times 256$ and RGB three channels. We are interested in generating
multiple texture images that are visually similar to it. By randomly
cropping texture patches of size $32 \times 32$ out of the source image,
we obtain a collection of texture samples, which serves as the input to
the NITES system. The dimension of these patches is $32 \times 32
\times 3=3072$. Their augmented vectors form source space $S_0$. The
NITES system is designed to generate texture patches of the same size
that are visually similar to samples in $S_0$.  This is feasible if we
can capture both global and local patterns of these samples. There are
two paths in Fig.~\ref{fig:framework}. The blue arrows go from left to
right, denoting the successive subspace embedding process. The red
arrows go from right to left, denoting the successive subspace
generation process.  We can generate as many texture patches as desired
using this procedure. In order to generate a texture image of a larger
size, we perform image quilting~\cite{efros2001image} based on
synthesized patches. 

\subsection{Successive Subspace Embedding (SSE)}\label{subsec:sse}

The global structure of an image (or an image patch) can be well
characterized by spectral analysis. Yet, it is weak in capturing local
detail such as boundaries between regions. Joint spatial-spectral
representations offer an ideal solution to the description of both
global shape and local detail information.  Embedding model $E_0^1$ find
a proper subspace, $S_1$, in $S_0$ while embedding model $E_1^2$ find a
proper subspace, $S_2$, in $S_1$. As shown in Fig.~\ref{fig:framework}
NITES applies two-stage transforms.  They correspond to $E_0^1$ and
$E_1^2$, respectively.  Specifically, we can apply the c/w Saab
transform in each stage to achieve the successive subspace embedding
(SSE) task.  In the following, we provide a brief review on the Saab
transform~\cite{kuo2019interpretable} and the c/w Saab
transform~\cite{chen2020pixelhopplus}. 

We partition each input patch into non-overlapping blocks, each of which
has a spatial resolution of $I_0 \times I_0$ with $K_0$ channels.  We
flatten 3D tensors into 1D vectors, and decompose each vector into the
sum of one DC and multiple AC spectral components. The DC filter is a
all-ones filter weighted by a constant. AC filters are obtained by
applying the principal component analysis (PCA) to DC-removed residual
tensor.  By setting $I_0=2$ and $K_0=3$, we have a tensor block of
dimension $2 \times 2 \times 3=12$.  Filter responses of PCA can be
positive or negative. There is a sign confusion problem
~\cite{kuo2016understanding, kuo2017cnn} if both of them are allowed to
enter the transform in the next stage. To avoid sign confusion, a
constant bias term is added to all filter responses to ensure that all
responses become positive - leading to the name of the "subspace
approximation with adjusted bias (Saab)" transform.  The Saab transform
is a date-driven transform, which is significantly different from
traditional transforms (e.g. Fourier and wavelet transforms) which are
data independent.  We partition AC channels into low- and high-frequency
two bands. The energy of high-frequency channels (shaded by gray color
in Fig.~\ref{fig:framework}) is low and they are discarded for dimension
reduction without affecting the performance much. The energy of
low-frequency channels (shaded by blue color in
Fig.~\ref{fig:framework}) is higher.  For a tensor of dimension 12, we
have one DC and 11 AC components. Typically, we select $K_1=6$ to 10
leading AC components and discard the rest. Thus, after $E_0^1$, one 12D
tensor becomes $K_1$-D vector, which is illustrated by dots in subspace
$S_1$.  The $K_1$-D response vectors are fed into the next stage for
another transform. 

The channel-wise (c/w) Saab transform \cite{chen2020pixelhopplus} exploits
the weak correlation property between channels so that the Saab
transform can be applied to each channel separately (see the middle part
of Fig.~\ref{fig:framework}). The c/w Saab transform offers an improved 
version of the standard Saab transform with a smaller model size. 

One typical setting used in our experiments is shown below.
\begin{itemize}
    \item Dimension of the input patch ($\tilde{D}_0$): $32\times32\times3=3072$;
    \item Dimension of subspace $\tilde{S}_1$ ($\tilde{D}_1$):
    $16\times16\times10=2560$ (by keeping 10 channels in Hop-1);
    \item Dimension of subspace $\tilde{S}_2$ ($\tilde{D}_2$):
    $8\times8\times27=1728$ (by keeping 27 channels in Hop-2).
\end{itemize}
Note that the ratio between $\tilde{D}_1$ and $\tilde{D}_0$ is 83.3\%
while that between $\tilde{D}_2$ and $\tilde{D}_1$ is 67.5\%. We are
able to reduce the dimension of the source space to that of the core
subspace by a factor of 56.3\%. In the reverse path indicated by red
arrows, we need to develop a multi-stage generation process. It should
also be emphasized that users can choose channel numbers in Hop-1 and
Hop-2 flexibly. Thus, NITES is a non-parametric method. 

The first-stage Saab transform provides the spectral information of the
nearest neighborhood, which is the first hop of the center pixel. By
generalizing from one to multiple hops, we can capture the information
in the short-, mid- and long-range neighborhoods. This is analogous to
increasingly larger receptive fields in deeper layers of CNNs. However,
filter weights in CNNs are learned from end-to-end optimization via
backpropagation while weights of the Saab filters in different hops are
determined by a sequence of PCAs in a feedforward unsupervised manner. 

\subsection{Core Subspace Generation (CSG)}\label{subsec:csg}

As stated before, we begin with the generation of samples in $S_n$,
which is accomplished by generation model $G_n$, in the reverse
generation path. In the current case, $n=2$. We need to characterize the
sample statistics in core subspace $S_2$ for the purpose of synthesis.
The statistics of a large texture image can be complicated.  We leverage
the semi-periodic property of texture so as to focus on patches of size
$32 \times 32$, whose statistics can be analyzed more easily.  

After the two-stage c/w Saab transform, the dimension of $S_2$ is
typically less than 2000. They are in form of c/w Saab coefficients. We
flatten these coefficients into a 1D vector, denoted by $\mathbf{z}$,
which is a random vector in $S_2$. To simplify the distribution
characterization of a high-dimensional random vector, we cluster
training samples into clusters and transform random vectors in each
cluster to a set of independent random variables. This is inspired by
the divide-and-conquer principle; namely, breaking down a difficult
problem into several easier sub-problems and solving them individually. 

We adopt a hierarchical K-Means clustering algorithm
\cite{nister2006scalable} to cluster the training samples of
$\mathbf{z}$ into $N$ clusters, which are denoted by $\{C_i\}$,
$i=0,\cdots,N-1$. Rather than modeling probability $P(\mathbf{z})$
directly, we model condition probability $P(\mathbf{z}\mid
\mathbf{z}\in C_i)$ with a fixed cluster index. The probability,
$P(\mathbf{z})$, can be written as
\begin{equation}\label{eq:prob}
P(\mathbf{z}) = \sum\limits_{i=0}^{N-1} P(\mathbf{z}\mid \mathbf{z}\in
C_i) \cdot P(\mathbf{z}\in C_i),
\end{equation}
where $P(\mathbf{z}\in C_i)$ is the percentage of data points in cluster
$C_i$. It is abbreviated as $p_i$, $i=0,\dots,N-1$ (see the right part of
Fig.~\ref{fig:framework}). 

Typically, a set of independent Gaussian random variables is used for
image generation. To do the same, we need to convert a collection of
correlated random vectors into a set of independent Gaussian random
variables. To achieve this objective, we transform random vector
$\mathbf{z}$ in cluster $C_i$ to a set of independent random variables
through independent component analysis (ICA), where non-Gaussianity
serves as an indicator for statistical independence. ICA finds
applications in noise reduction~\cite{hyvarinen1999sparse}, face
recognition~\cite{bartlett2002face}, and image
infusion~\cite{mitianoudis2007pixel}.  

Our implementation is detailed below.
\begin{enumerate}
    \item Apply PCA to $\mathbf{z}$ in cluster $C_i$ for dimension reduction and data whitening. 
    \item Apply FastICA~\cite{hyvarinen2000independent}, which is conceptually simple, computational efficient and robust to outliers, to the PCA output.  
    \item Compute the cumulative density function (CDF) of each ICA
    component of random vector $\mathbf{z}$ in each cluster based on its
    histogram of training samples. 
    \item Match the CDF in Step 3 with the CDF of a Gaussian random variable
    (see the right part of Fig.  \ref{fig:framework}), where the inverse CDF
    is obtained by resampling between bins with linear interpolation.  To
    reduce the model size, we quantize N-dimensional CDFs, which have $N$
    bins, with vector quantization (VQ) and store the codebook of quantized CDFs. 
\end{enumerate}

We encode $P(\mathbf{z}\in C_i)$ in Eq. (\ref{eq:prob}) to be the length
of a segment in the unit interval, $[0,1]$. All segments are
concatenated in order to build the unit interval. The segment index is
the cluster index. These segments is called the interval representation.
To draw a sample from subspace $S_2$, we use the uniform random number
generator to select a random number from interval $[0,1]$ to decide the
cluster index using the interval representation. 

To generate a new sample in $S_2$, we perform the following steps:
\begin{enumerate}
    \item Select a random number from the uniform random number generator to decide the cluster index.
    \item Draw a set of sample independently from the Gaussian distribution.
    \item Conduct histogram matching on the generated Gaussian samples using the inverse CDFs in the chosen cluster.
    \item Repeat Steps 1-3 if the generated sample of Step 3 has no value larger than a pre-set threshold.
    \item Perform the inverse transform of ICA and the inverse transform of PCA.
    \item Reshape the 1D vector into a 3D tensor and this tensor is the generated sample in $S_2$.
\end{enumerate}

\subsection{Successive Subspace Generation (SSG)}\label{subsec:ssg}

In this section, we examine generation model $G_{i+1}^i$, whose role is
to generate a sample in $S_i$ given a generated sample in $S_{i+1}$.
The core subspace generation process can generate a sample that captures
the global structure of an image patch but lacks in local detail such as
boundaries between regions. Embedding model, $E_i^{i+1}$, provides a
good description of local detail, which is obtained by the c/w Saab
transform. Thus, in the generation process, we conduct the inverse c/w
Saab transform on the generated sample in $S_{i+1}$. 

It is worthwhile to point out that both Saab and c/w Saab transforms
have the forward and the inverse transforms. If no high-frequency
channels are removed, they are lossless transforms. If some
high-frequency channels are removed, they become lossy transforms.  In
the current case, we use lossy transforms for dimension reduction in the
successive subspace embedding process and lossy inverse transforms in
the successive subspace generation process.  The inverse c/w Saab
transform is a deterministic one once the forward transform is
determined. It can be used for signal reconstruction. 

\begin{figure}[tb]
\begin{center}
\includegraphics[width=0.7\linewidth]{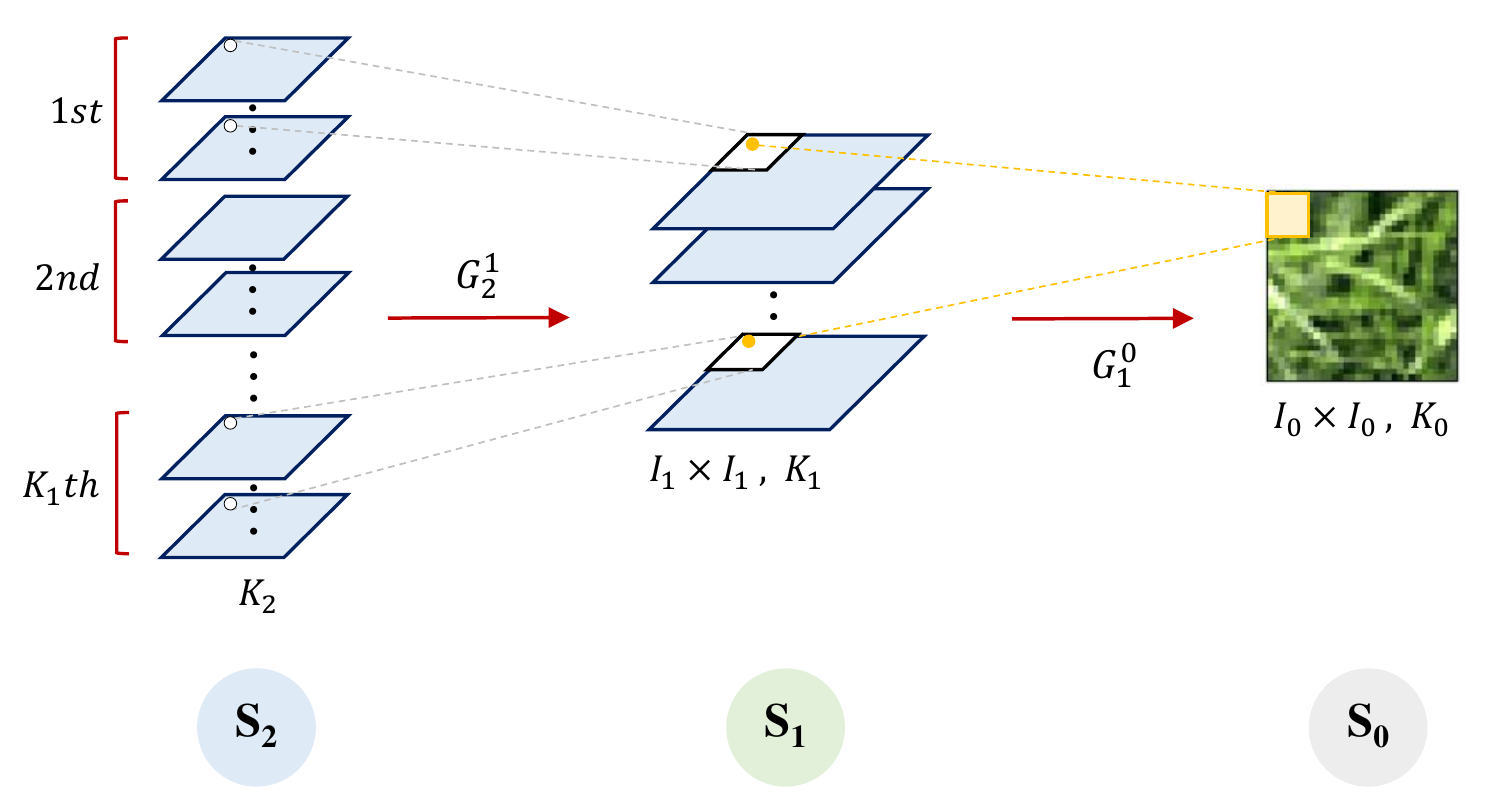}
\end{center}
\caption{Illustration of generation model $G_{i+1}^i$, which converts the
same spatial location of $K_{i+1}$ channels in subspace $S_{i+1}$ to a block of
size $I_i \times I_i$ of $K_i$ channels in space/subspace $S_i$.}\label{fig:ssg}
\end{figure}

In the following, we take generation model $G_2^1$ from $S_2$ and to
$S_1$ as an example to explain the generation process.  A generated
sample in $S_2$ can be partitioned into $K_1$ groups as shown in the
left part of Fig.~\ref{fig:ssg}. Each group of channels is composed by
one DC channel and several low-frequency AC channels. The $k$th group of
channels in $S_2$, whose number is denoted by $K_2^{(k)}$, is derived
from the $k$th channel in $S_1$. We apply the inverse c/w Saab transform
to each group individually. The inverse c/w Saab transform converts the
tensor at the same spatial location across $K_2^{(k)}$ channels
(represented by white dots in Fig.~\ref{fig:ssg}) in $S_2$ into a block
of size $I_i \times I_i$ (represented by the white parallelogram in
Fig.~\ref{fig:ssg}) in $S_1$, using the DC and AC components obtained in
the successive subspace embedding process.  After the inverse c/w Saab
transform, the Saab coefficients in $S_1$ form a generated sample in
$S_1$. The same procedure is repeated between $S_1$ and $S_0$. 

\begin{figure*}[tb]
\begin{center}
\includegraphics[width=0.7\linewidth]{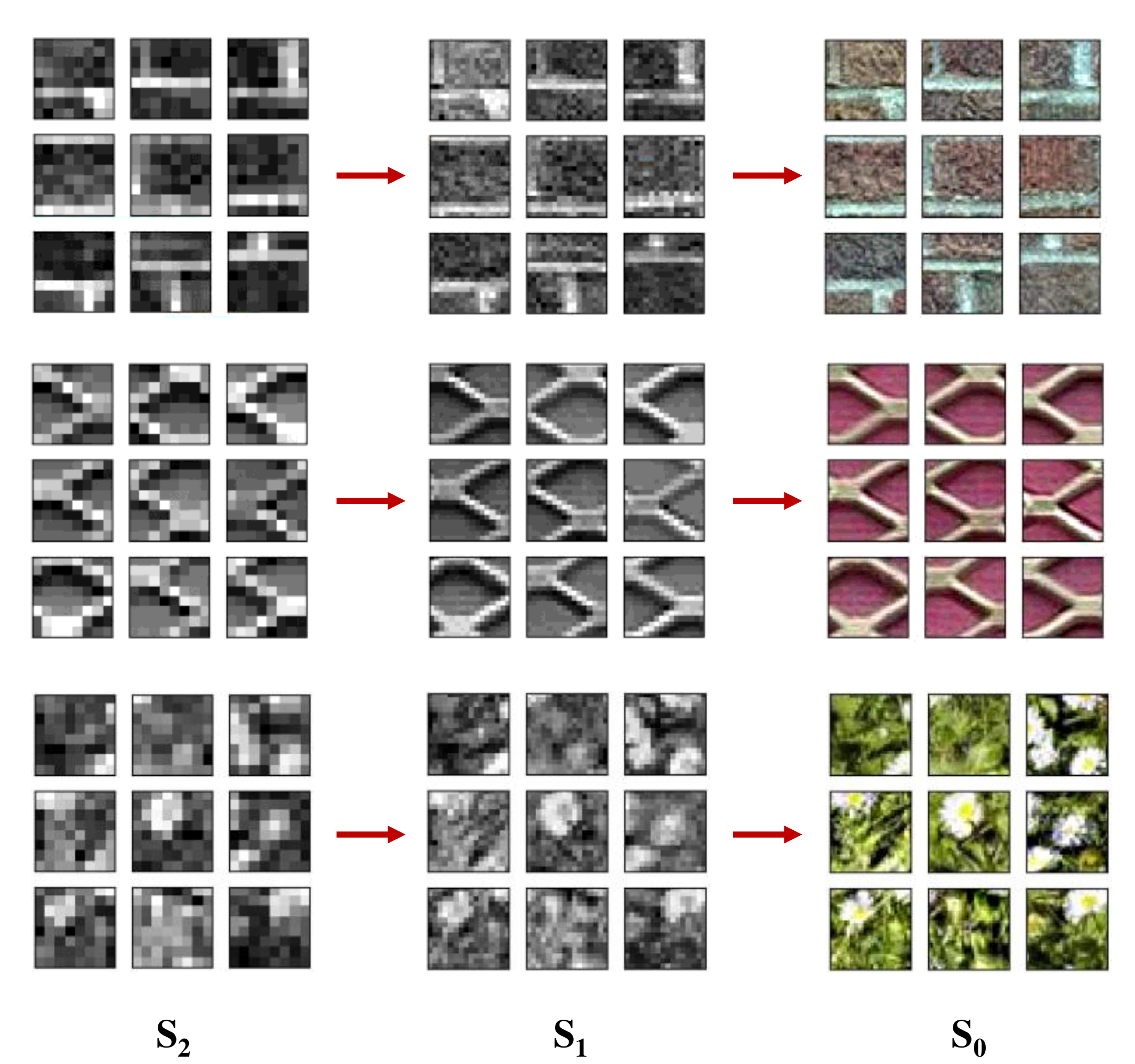}
\end{center}
\caption{Examples of several generated textures in core subspace $S_0$, 
intermediate subspace $S_1$ and source space $S_0$.}\label{fig:between}
\end{figure*}

Examples of several generated textures in core subspace $S_0$,
intermediate subspace $S_1$ and source space $S_0$ are shown in
Fig.~\ref{fig:between}, where we display the generated DC channel of
group 1 (the first channel in Fig.~\ref{fig:ssg}) in $S_2$ and $S_1$ and
the generated texture patch in $S_0$. These DC channels offer gray-scale
low-resolution patterns of a generated sample. NITES can generate
samples containing more local detail gradually. 

\section{Experiments}\label{sec:experiments}

\subsection{Experimental Setup}

The following hyper parameters (see Fig. \ref{fig:framework}) are used in our experiments.
\begin{itemize}
    \item Input filter window size to Hop-1: $I_0=2$,
    \item Input filter window size to Hop-2: $I_1=2$,
    \item Selected channel numbers in Hop-1 ($K_1$): $6\sim10$,
    \item Selected channel numbers in Hop-2 ($K_2$): $20\sim30$.
\end{itemize}
The analysis filter window size is the same as the generation window
size. All windows are non-overlapping with each other. The actual
channel numbers $K_1$ and $K_2$ are texture dependent. That is, we
examine the energy distribution based on the PCA eigenvalue plot and
choose the knee point where the energy becomes nearly flat.

\subsection{An Example: Brick Wall Texture Generation}

\begin{figure*}[htb]
\begin{center}
\includegraphics[width=0.9\linewidth]{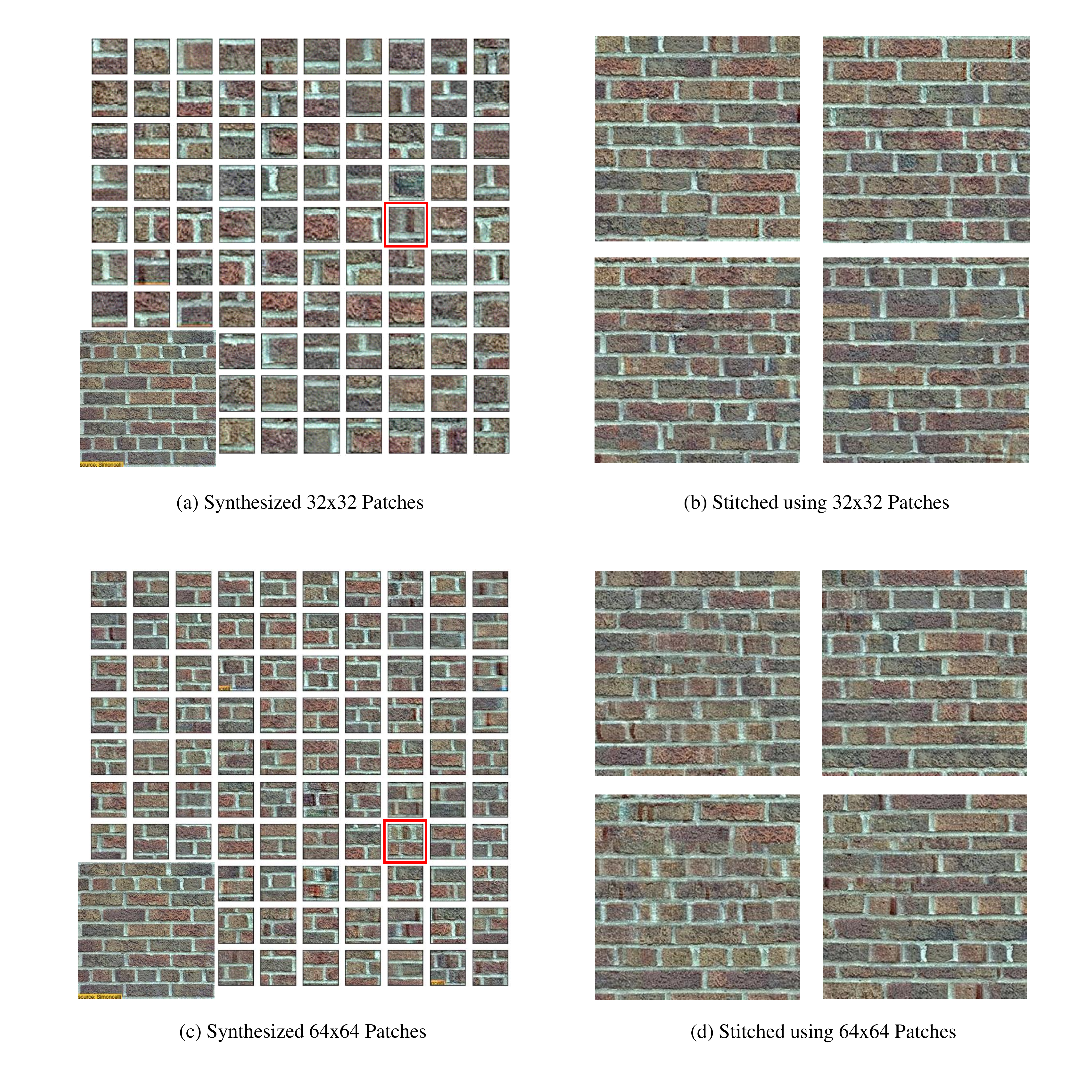}
\end{center}
\caption{Examples of generated $brick\_wall$ texture patches and
stitched images of larger sizes, where the image in the bottom-left
corner is the exemplary texture image and the patches highlighted by red
squared boxes are unseen patterns.}\label{fig:synpatch}
\end{figure*}

We show generated \emph{brick\_wall} texture patches of size
$32\times32$ and $64\times64$ in Figs.~\ref{fig:synpatch}(a) and (c).
We performed two-stage c/w Saab transforms on $32\times32$ patches and
three-stage c/w Saab transforms on $64\times64$ patches, whose core
subspace dimensions are equal to 1728 and 4032, respectively. Patches in
these figures were synthesized by running the NITES method in one
hundred rounds. Randomness in each round primarily comes from two
factors: 1) random cluster selection, and 2) random seed vector
generation. 

Generated patches retain the basic shape of bricks and the diversity of
brick texture. We observe some unseen patterns generated by NITES, which
are highlighted by red squared boxes in Figs.~\ref{fig:synpatch} (a) and
(c). As compared with generated $32\times32$ patches, generated
$64\times64$ patches could be blurry sometimes (e.g., the one in the
upper right corner) due to a higher source dimension. 

\begin{figure*}[htb]
\begin{center}
\includegraphics[width=0.9\linewidth]{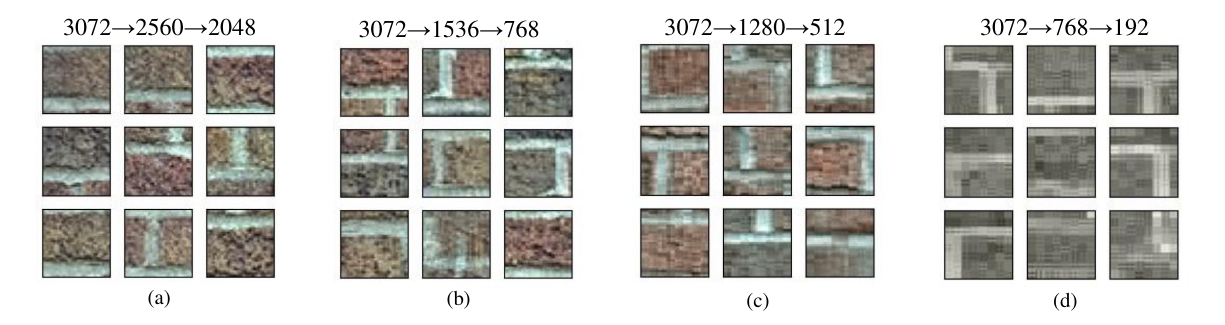}
\end{center}
\caption{An illustration of how dimension growth affects the generation
results. The numbers above the figure indicates the dimensions of $S_0$,
$S_1$ and $S_2$, respectively.}\label{fig:dimanalysis}
\end{figure*}

As a non-parametric generation model, NITES has freedom in choosing
multiple settings under the same pipeline. For example, it can select
different channel numbers in $\tilde{S}_1$ and $\tilde{S}_2$ to derive
different generation results. Four settings are listed in
Table~\ref{table:settings}. The corresponding generation results are
shown in Fig.~\ref{fig:dimanalysis}. Dimensions decrease faster from
(a) to (d). The quality of generated results becomes poorer due to 
smaller dimensions of the core subspace, $\tilde{S}_2$, and the
intermediate subspace, $\tilde{S}_1$. 

\begin{table}[h]
\begin{center}
\caption{The settings of our four generation model.} \label{table:settings}
\begin{tabular}{cccc} \\ \hline
Setting & $\tilde{D}_0$ & $\tilde{D}_1$ & $\tilde{D}_2$ \\ \hline
a       & 3072          & 2560          & 2048 \\ 
b       & 3072          & 1536          & 768 \\ 
c       & 3072          & 1280          & 512 \\ 
d       & 3072          & 768           & 192 \\ \hline
\end{tabular}
\end{center}
\end{table}

To generate larger texture images, we first generate 5,000 texture
patches and perform image quilting~\cite{efros2001image} with them. The
results after quilting are shown in Figs.~\ref{fig:synpatch} (b) and
(d).  All eight images are of the same size, $256\times256$. They are
obtained using different initial patches for the image quilting process.
By comparing the two sets of stitched images, the global structure of
the brick wall is better preserved using larger patches (i.e. of size
$64 \times 64$) while its local detail is a little bit blurry sometimes. 

\subsection{Performance Benchmarking with CNN-based Methods}

\begin{figure*}[htb]
\begin{center}
\includegraphics[width=\textwidth, height=0.92\textheight, keepaspectratio]{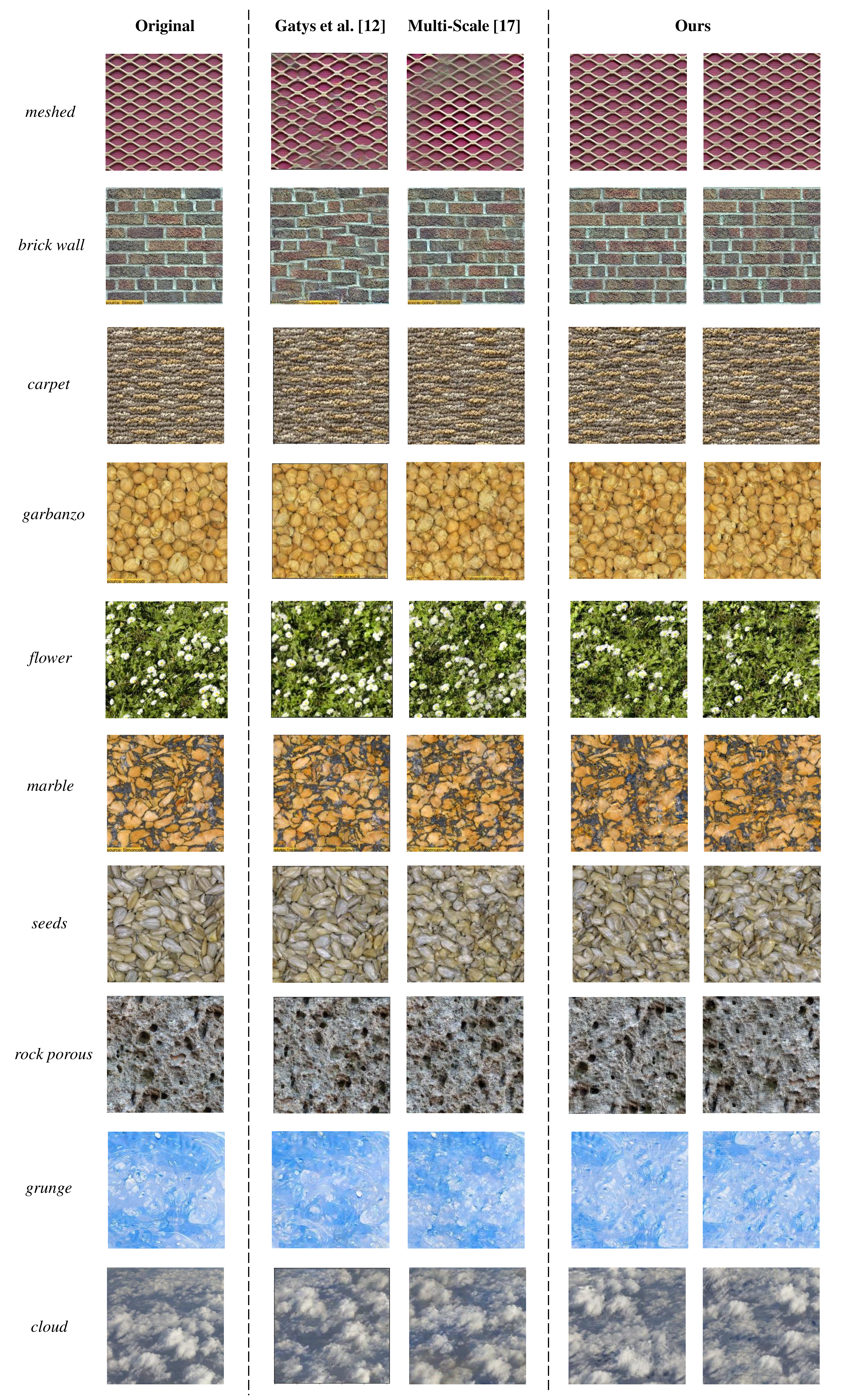}
\end{center}
\caption{Comparison of synthesized texture images using two CNN-based
methods and NITES (from left to right): exemplary texture images,
texture images generated by \cite{gatys2015texture} texture images
generated by \cite{ ustyuzhaninov2017does} and two texture images
generated by NITES.} \label{fig:synimg}
\end{figure*}

{\bf Visual Quality Comparison.} The quality of synthesized texture is
usually evaluated by human eyes. A diversity loss function was proposed
to measure texture diversity for CNN-based methods~\cite{shi2020fast, li2017diversified}. Yet, NITES dose not have a loss function.  Thus, we
show synthesis results of two CNN methods and NITES side by side in
Fig.~\ref{fig:synimg}.  Exemplary texture images are collected from
those in~\cite{gatys2015texture, ustyuzhaninov2017does, portilla2000parametric} or the Internet for the illustration purpose. The two benchmarking CNN methods were
proposed by Gatys {\em et al.}~\cite{gatys2015texture} and Ustyuzhaninov
{\em et al.} \cite{ustyuzhaninov2017does}. We run the codes provided by
them, and show their results in the second and third columns of
Fig.~\ref{fig:synimg}, respectively. We used the default setting of the iteration number, which is 2000 in~\cite{gatys2015texture} and 4000 in~\cite{gatys2015texture} for visualization. Two results generated by NITES are
shown in the last two columns. They are obtained in two different runs.
For texture \emph{meshed}, we see the brown fog artifact in
~\cite{gatys2015texture, ustyuzhaninov2017does}, which is visually
apparent. However, it does not show up in our generation.  As
demonstrated by these examples, NITES can generate high quality and
visually pleasant texture images. 

{\bf Comparison of Texture Generation Time.} It is worthwhile to compare
texture synthesis time of different texture image generation methods.
The results are shown in Table~\ref{table:time}. All experiments were
conducted in the same machine composed by 12 CPUs (Intel Core i7-5930K
CPU at 3.50GHz) and 1 GPU (GeForce GTX TITAN X).  No GPU was needed in
our model but it was needed in~\cite{gatys2015texture, ustyuzhaninov2017does}. We set the iteration number to 1000
for~\cite{gatys2015texture} and 100 for~\cite{ustyuzhaninov2017does}.
NITES generated 5,000 $32\times32$ texture patches, 81 of which were
stitched into the final texture image.  For all three methods, we
measure the time needed in generating one image of size $256\times 256$.
As shown in Table~\ref{table:time}, NITES generates one texture image
with 213.04 seconds while Gatys' method and Ustyuzhaninov's method
demand 513.98 and 949.64 seconds, respectively.  NITES is significantly
faster. 

\begin{table}[htb]
\begin{center}
\caption{Comparison of time needed to generate one texture image.} \label{table:time}
\begin{tabular}{cc} \\ \hline
Methods                                  &  Time (seconds) \\ \hline
Gatys et al.~\cite{gatys2015texture}     &  513.98  \\ 
Multi-Scale~\cite{ustyuzhaninov2017does} &  949.64  \\
NITES (Ours)                             &  213.04  \\ \hline
\end{tabular}
\end{center}
\end{table}

{\bf Breakdown of NITES' Generation Time.} We can break down the texture
generation time of NITES into three parts: 1) successive subspace
embedding (i.e., the forward embedding path), 2) core and successive
subspace generations (i.e., the reverse generation path) and 3) the
quilting process. Time required by each part is shown in
Table~\ref{table:processtime}. They demand 24.24, 108.03 and 8.08
seconds, respectively. To generate multiple texture images from the same
exemplary texture, we need to run the first part once but the second and
third parts multiple times (one run for one new synthesis).  Thus, it is
fair to focus on the last two parts only for single texture image
generation, which is equal to 116 seconds.  In contrast, the two
benchmarking CNN methods do not have such a breakdown.  They need to go
through the whole pipeline to generate one new texture image. 

\begin{table}[htb]
\begin{center}
\caption{The time of three processes in our method.}\label{table:processtime}
\begin{tabular}{cc} \\ \hline
Process           &   Time (seconds)  \\ \hline
Forware Embedding &   24.24  \\
Reverse Generation&  108.03 \\
Quilting          &    8.08 \\  \hline
\end{tabular}
\end{center}
\end{table}

\section{Conclusion and Future Work}\label{sec:conclusion}

A non-parametric interpretable texture synthesis (NITES) method was
proposed based on a new texture analysis and synthesis framework in this
work. Texture can be analyzed and represented effectively using the
multi-stage c/w Saab transforms that offer a sequence of joint
spatial-spectral representations. The sample distribution in the core
subspace was carefully studied, which allows us to build a core subspace
generation model. Furthermore, a successive subspace generation model
was developed to build a higher-dimensional subspace based on a
lower-dimensional subspace. As a result, new texture samples can be
generated by mimicking probabilities and/or conditional probabilities of
source texture patches. Extensive experimental results were conducted to
demonstrate the power of the proposed NITES method.  It can generate
visually pleasant texture images effectively, including some unseen
patterns. 

As to future research extension, there are several directions worth
exploration. It appears to be an important topic in controlling the
growth of subspace dimensions in the generation process. Is it
beneficial to introduce more intermediate subspaces between the source
and the core? Can we apply the same model for other image generation
such as human faces, digits, scenes and objects? Is it possible to
generalize the framework to image inpainting? How to compare our
generation model and GAN? These are all open and interesting questions
for further investigation. 

\section*{Acknowledgment}

This work was supported by MediaTek. Computation for the work was supported by the University of Southern
California's Center for High Performance Computing (hpc.usc.edu).

\bibliographystyle{unsrt}  
\bibliography{paper.bib}

\end{document}